\documentclass[runningheads]{llncs}

\usepackage{graphicx}
\usepackage{amsmath,amssymb} 
\usepackage{color}
\usepackage[width=122mm,left=12mm,paperwidth=146mm,height=193mm,top=12mm,paperheight=217mm]{geometry}

\usepackage{caption}
\usepackage[hidelinks]{hyperref} 
\hypersetup{
	colorlinks=true,
	anchorcolor=black,
	citecolor=green,
	linkcolor=red,
	filecolor=cyan,      
	urlcolor=magenta
}
\usepackage[hyperpageref]{backref} 
\usepackage{multicol}
\usepackage{makecell}
\usepackage{amsmath}
\usepackage{amssymb}
\usepackage{rotating}
\usepackage{pdflscape}
\usepackage{mathrsfs}
\usepackage{cleveref}
\usepackage{subcaption}
\usepackage{array,multirow}
\usepackage{color}
\usepackage{wrapfig}
\usepackage[innercaption]{sidecap}
\begin{document}
%
\pagestyle{headings}
\mainmatter

\newcommand{\todo}[1]{{\color{red}#1}}



\title{Deep Virtual Stereo Odometry:\\Leveraging Deep Depth Prediction for 
Monocular Direct Sparse Odometry} 

\titlerunning{DVSO: Leveraging Deep Depth Prediction 
for Monocular DSO}

\author{Nan Yang\inst{1,2} \and Rui Wang\inst{1,2} \and J\"org 
St\"uckler\inst{1} \and Daniel Cremers\inst{1,2}}
\institute{Technical University of Munich \and Artisense \\
\email{\{yangn,wangr,stueckle,cremers\}@in.tum.de}
}

\authorrunning{N. Yang, R. Wang, J. St\"uckler and D. Cremers}


\maketitle


\begin{abstract}

Monocular visual odometry approaches that purely rely on geometric cues are prone to scale drift and require sufficient motion parallax in successive frames for motion estimation and 3D reconstruction.
In this paper, we propose to leverage deep monocular depth prediction to overcome limitations of geometry-based monocular visual odometry.
To this end, we incorporate deep depth predictions into Direct Sparse Odometry 
(DSO) 
as direct virtual stereo measurements.
For depth prediction, we design a novel deep network that refines predicted depth from a single image in a two-stage process.
We train our network in a semi-supervised way on photoconsistency in stereo images and on consistency with accurate sparse depth reconstructions from Stereo DSO.
Our deep predictions excel state-of-the-art approaches for monocular depth on the KITTI benchmark.
Moreover, our Deep Virtual Stereo Odometry clearly exceeds previous monocular and deep-learning based methods in accuracy.
It even achieves comparable performance to the state-of-the-art stereo methods, while only relying on a single camera.



\end{abstract}

\section{Introduction}\label{sec:introduction}

Visual odometry (VO) is a highly active field of research in computer vision with a plethora of applications in domains such as autonomous driving, robotics, and augmented reality.
VO with a single camera using traditional geometric approaches inherently suffers from the fact that camera trajectory and map can only be estimated up to an unknown scale which also leads to scale drift. 
Moreover, sufficient motion parallax is required to estimate motion and structure from successive frames. 
To avoid these issues, typically more complex sensors such as active depth cameras or stereo rigs are employed. 
However, these sensors require larger efforts in calibration and increase the costs of the vision system.

Metric depth can also be recovered from a single image if a-priori knowledge about the typical sizes or appearances of objects is used. 
Deep learning based approaches tackle this by training deep neural networks on large amounts of data. 
In this paper, we propose a novel approach to monocular visual odometry, Deep Virtual Stereo Odometry (DVSO), which incorporates deep depth predictions into a geometric monocular odometry pipeline.
We use deep stereo disparity for virtual direct image alignment constraints within a framework for windowed direct bundle adjustment (e.g. Direct Sparse Odometry~\cite{engel2017direct}).
DVSO achieves comparable performance to the state-of-the-art stereo visual odometry systems on the KITTI odometry benchmark.
It can even outperform the state-of-the-art geometric VO methods when tuning scale-dependent parameters such as the virtual stereo baseline. 

As an additional contribution, we propose a novel stacked residual network architecture that refines disparity estimates in two stages and is trained in a semi-supervised way.
In typical supervised learning approaches~\cite{eigen2014depth,li2015depth,laina2016deeper}, depth ground truth needs to be
acquired for training with active sensors like RGB-D cameras and 3D laser 
scanners which are costly to obtain.
Requiring a large amount of such labeled data is an additional burden that limits generalization to new environments. 
Self-supervised~\cite{garg2016unsupervised,godard2016unsupervised} and 
unsupervised learning 
approaches~\cite{zhou2017unsupervised}, on the other hand, overcome this 
limitation and do not require additional active sensors. 
Commonly, they train the networks on photometric consistency, for example in stereo imagery~\cite{garg2016unsupervised,godard2016unsupervised}, which reduces the effort for 
collecting training data.
Still, the current self-supervised approaches are not as accurate as supervised 
methods~\cite{kuznietsov2017semi}.
We combine self-supervised and supervised training, but avoid the costly collection of LiDAR data in our approach.
Instead, we make use of Stereo Direct Sparse Odometry (Stereo DSO~\cite{wang2017stereoDSO}) to provide accurate sparse 3D reconstructions on the training set.
Our deep depth prediction network outperforms the current state-of-the-art 
methods on KITTI.

A video demonstrating our methods as well as the results is available at 
\url{https://youtu.be/sLZOeC9z_tw}.
\begin{figure}[t]
	\includegraphics[width=\textwidth]{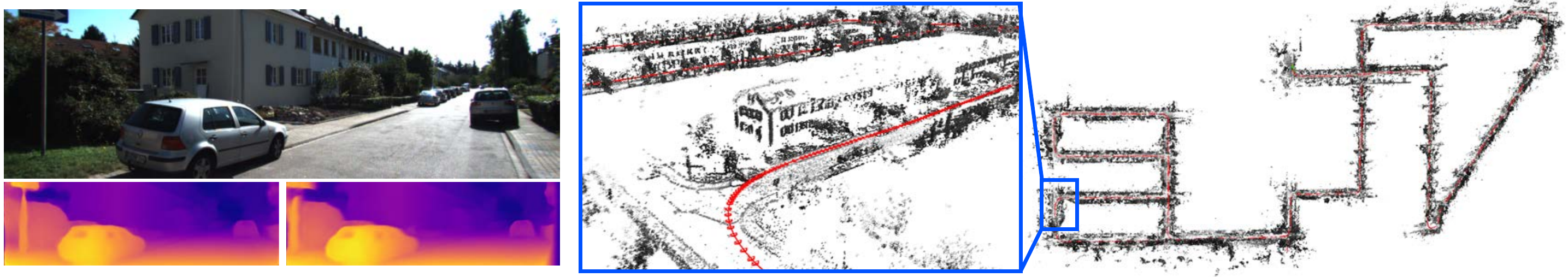}
	\caption{DVSO achieves monocular visual odometry on KITTI on par with 
	state-of-the-art stereo methods. It uses deep-learning based left-right 
	disparity predictions (lower left) for initialization and virtual stereo 
	constraints in an optimization-based direct visual odometry pipeline. This 
	allows for recovering accurate metric estimates.}
\end{figure}

\subsection{Related Work}

\textbf{Deep learning for monocular depth estimation.} 
Deep learning based approaches have recently achieved great advances in monocular depth estimation.
Employing deep neural network avoids the hand-crafted features used in previous methods~\cite{saxena2006learning,hoiem2005automatic}. 
Supervised deep learning~\cite{eigen2014depth,li2015depth,laina2016deeper} has recently shown great success for monocular depth estimation.
Eigen et al.~\cite{eigen2014depth,eigen2015predicting} propose a two scale CNN architecture which directly predicts the depth map from a single image. 
Laina et al.~\cite{laina2016deeper} propose a residual 
network~\cite{he2016deep} based fully convolutional encoder-decoder 
architecture~\cite{long2015fully} with a robust regression loss function. 
The aforementioned supervised learning approaches need large amounts of ground-truth depth data for training. 
Self-supervised approaches~\cite{garg2016unsupervised,xie2016deep3d,godard2016unsupervised} overcome this limitation by exploiting photoconsistency and geometric constraints to define loss functions, for example, in a stereo camera setup.
This way, only stereo images are needed for training which are typically easier to obtain than accurate depth measurements from active sensors such as 3D lasers or RGB-D cameras. 
Godard et al.~\cite{godard2016unsupervised} achieve the state-of-the-art depth estimation accuracy for a fully self-supervised approach.
The semi-supervised scheme proposed by Kuznietsov et al.~\cite{kuznietsov2017semi} combines the self-supervised loss with supervision with sparse LiDAR ground truth. 
They do not need multi-scale depth supervision or left-right consistency in their loss, and achieve better performance than the self-supervised approach in~\cite{godard2016unsupervised}. The limitation of this semi-supervised approach is the requirement for LiDAR data which are costly to collect.
In our approach we use Stereo Direct Sparse Odometry to obtain sparse depth ground-truth for semi-supervised training.
Since the extracted depth maps are even sparser than LiDAR data, we also employ multi-scale self-supervised training and left-right consistency as in Godard et al.~\cite{godard2016unsupervised}.
Inspired by~\cite{ilg2017flownet,pang2017cascade}, we design a stacked network 
architecture leveraging the concept of residual learning~\cite{he2016deep}.

\textbf{Deep learning for VO / SLAM.} 
In recent years, large progress has been achieved in the development of monocular VO and SLAM methods~\cite{mur2017orb,engel2014lsd,engel2017direct,newcombe2011dtam}.
Due to projective geometry, metric scale cannot be observed with a single camera~\cite{strasdat2010scale} which introduces scale drift. 
A popular approach is hence to use stereo cameras for VO~\cite{engel2015large,engel2017direct,mur2017orb} which avoid scale ambiguity and leverage stereo matching with a fixed baseline for estimating 3D structure. 
While stereo VO delivers more reliable depth estimation, it requires self-calibration for long-term operation~\cite{dang2009continuous,yin2017scale}. The integration of a second camera also introduces additional costs.
Some recent monocular VO approaches have integrated monocular depth estimation~\cite{yin2017scale,tateno2017cnn} to recover the metric scale by scale-matching.
CNN-SLAM~\cite{tateno2017cnn} extends LSD-SLAM~\cite{engel2014lsd} by predicting depth with a CNN and refining the depth maps using Bayesian filtering~\cite{engel2014lsd,engel2013iccv}. 
Their method shows superior performance over monocular 
SLAM~\cite{engel2014lsd,mur2015orb,yang2018challenges,pizzoli2014remode} on 
indoor datasets~\cite{handa2014benchmark,sturm2012benchmark}. Yin et 
al.~\cite{yin2017scale} propose to use convolutional neural fields and 
consecutive frames to improve the monocular depth estimation from a CNN. 
Camera motion is estimated using the refined depth. 
CodeSLAM~\cite{bloesch2018codeslam} focuses on the challenge of dense 3D reconstruction. 
It jointly optimizes a learned compact representation of the dense geometry with camera poses. 
Our work tackles the problem of odometry with monocular cameras and integrates deep depth prediction with multi-view stereo to improve camera pose estimation.
Another line of research trains networks to directly predict the ego-motion end-to-end using supervised~\cite{wang2017deepvo} or unsupervised learning~\cite{zhou2017unsupervised,li2017undeepvo}.
However, the estimated ego-motion of these methods is still by far inferior to geometric visual odometry approaches.
In our approach, we phrase visual odometry as a geometric optimization problem but incorporate photoconsistency constraints with state-of-the-art deep monocular depth predictions into the optimization.
This way, we obtain a highly accurate monocular visual odometry that is not prone to scale drift and achieves comparable results to traditional stereo VO methods.


\section{Semi-Supervised Deep Monocular Depth Estimation}\label{sec:depth_est}

In this section, we will introduce our semi-supervised approach to deep monocular depth estimation.
It builds on three key ingredients: self-supervised learning from photoconsistency in a stereo setup similar to~\cite{godard2016unsupervised},
supervised learning based on accurate sparse depth reconstruction by Stereo DSO, and two-stage refinement of the network predictions in a stacked encoder-decoder architecture.


\begin{figure}[tb]
	\centering
	\includegraphics[width=\textwidth]{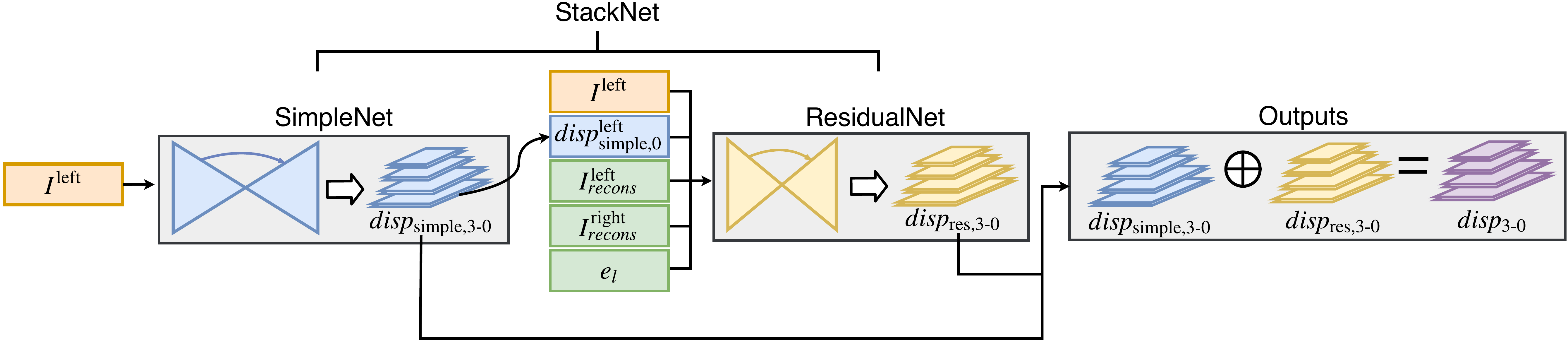}
	\caption{Overview of StackNet architecture.}
	\label{fig:structure1}
\end{figure}

\subsection{Network Architecture}
We coin our architecture StackNet since it stacks two sub-networks, SimpleNet and ResidualNet, as depicted in~\Cref{fig:structure1}. 
Both sub-networks are fully convolutional deep neural network adopted from DispNet~\cite{mayer2016large} 
with an encoder-decoder scheme. ResidualNet has fewer layers and takes the outputs of SimpleNet as inputs. 
Its purpose is to refine the disparity maps predicted by SimpleNet by learning an additive residual signal.
Similar residual learning architectures have been successfully applied to 
related deep learning tasks~\cite{ilg2017flownet,pang2017cascade}. The detailed 
network architecture is illustrated in the 
\href{https://vision.in.tum.de/_media/spezial/bib/yang2018dvso-supp.pdf}{supplementary
 material}.

\textbf{SimpleNet.} 
SimpleNet is an encoder-decoder architecture with a ResNet-50 based encoder and 
skip connections between corresponding encoder and decoder layers.
The decoder upprojects the feature maps to the original 
resolution and generates 4 pairs of disparity maps $\mathit{disp}_{\mathit{simple},s}^{\mathit{left}}$ and $\mathit{disp}_{\mathit{simple},s}^{\mathit{right}}$ in different resolutions $s \in [0,3]$. 
The upprojection is implemented by resize-convolution~\cite{odena2016deconvolution}, i.e. a nearest-neighbor upsampling layer by a factor of two followed by a convolutional layer. 
The usage of skip connections enables the decoder to recover high-resolution results with fine-grained details.

\textbf{ResidualNet.} 
The purpose of ResidualNet is to further refine the disparity maps predicted by SimpleNet. 
ResidualNet learns the residual signals $\mathit{disp}_{\mathit{res},s}$ to the disparity maps $\mathit{disp}_{\mathit{simple},s}$ (both left and right and for all resolutions). 
Inspired by FlowNet 2.0~\cite{ilg2017flownet}, the inputs to ResidualNet contain various information on the prediction and the errors made by SimpleNet: 
we input $I^{\mathit{left}}$, 
$\mathit{disp}_{\mathit{simple},0}^{\mathit{left}}$, 
$I^{\mathit{right}}_{\mathit{recons}}$, 
$I^{\mathit{left}}_{\mathit{recons}}$ and $e_l$, where
\begin{itemize}
	\item $I^{\mathit{right}}_{\mathit{recons}}$ is the reconstructed right image by warping 
	$I^{\mathit{left}}$ using $\mathit{disp}_{\mathit{simple},0}^{\mathit{right}}$.
	\item $I^{\mathit{left}}_{\mathit{recons}}$ is the generated left image by back-warping 
	$I^{\mathit{right}}_{\mathit{recons}}$ using $\mathit{disp}_{\mathit{simple},0}^{\mathit{left}}$.
	\item $e_l$ is the $\ell1$ reconstruction error between $I^{\mathit{left}}$ and 
	$I^{\mathit{left}}_{\mathit{recons}}$
\end{itemize}
For the warping, rectified stereo images are required while stereo camera intrinsics and extrinsics are not needed as our network directly outputs disparity.

The final refined outputs $\mathit{disp}_s$ are
	$\mathit{disp}_s = \mathit{disp}_{\mathit{simple},s} \oplus \mathit{disp}_{res,s}, s \in [0, 3]$,
where $\oplus$ is element-wise summation. 
The encoder of ResidualNet contains 12 residual blocks in total and predicts 4 scales of residual 
disparity maps as SimpleNet. Adding more layers does not further improve performance in our experiments.
Notably, only the left image is used as an input to either SimpleNet and ResidualNet, while the right image is not required.
However, the network outputs a refined disparity map for the left and right stereo image.
Both facts will be important for our monocular visual odometry approach.

\subsection{Loss Function}
We define a loss $\mathcal{L}_s$ at each output scale $s$, resulting in the total loss
$\mathcal{L} = \sum_{s=0}^{3}\mathcal{L}_s.$
The loss at each scale $\mathcal{L}_s$ is a linear combination of five terms which are symmetric in left and right images,
\begin{equation}
\begin{split}
\mathcal{L}_s
& = \alpha_{U}\left(\mathcal{L}^{\mathit{left}}_{U} + \mathcal{L}^{\mathit{right}}_{U}\right) + 
\alpha_{S}\left(\mathcal{L}^{\mathit{left}}_{S} + \mathcal{L}^{\mathit{right}}_{S}\right) + 
\alpha_{lr}\left(\mathcal{L}^{\mathit{left}}_{\mathit{lr}} + \mathcal{L}^{\mathit{right}}_{\mathit{lr}}\right) \\
& + \alpha_{\mathit{smooth}}\left(\mathcal{L}^{\mathit{left}}_{\mathit{smooth}} + 
\mathcal{L}^{\mathit{right}}_{\mathit{smooth}}\right)  + \alpha_{\mathit{occ}}\left(\mathcal{L}^{\mathit{left}}_{\mathit{occ}} + 
\mathcal{L}^{\mathit{right}}_{\mathit{occ}}\right),
\end{split}
\label{eq:total_loss_dl}
\end{equation}
where $\mathcal{L}_{U}$ is a self-supervised loss, $\mathcal{L}_{S}$ is a 
supervised loss, $\mathcal{L}_{\mathit{lr}}$ is a left-right consistency loss, 
$\mathcal{L}_{\mathit{smooth}}$ is a smoothness term encouraging the predicted 
disparities to be locally smooth and $\mathcal{L}_{\mathit{occ}}$ is an occlusion 
regularization term.
In the following, we detail the left components $\mathcal{L}^{\mathit{left}}$ of the 
loss function at each 
scale. The right components $\mathcal{L}^{\mathit{right}}$ are defined 
symmetrically.

\textbf{Self-supervised loss.} 
The self-supervised loss measures the quality of the reconstructed images. The 
reconstructed image is generated by warping the input image into the view of the other rectified stereo image. 
This procedure is fully (sub-)differentiable for bilinear sampling~\cite{jaderberg2015spatial}. 
Inspired by~\cite{godard2016unsupervised,zhao2015l2}, the quality of the 
reconstructed image is measured with the combination of the $\ell_1$ loss and 
single scale structural similarity 
(SSIM)~\cite{wang2004image}:  
\begin{equation}
\label{eq:unsup_loss}
\begin{split}
\mathcal{L}_U^{\mathit{left}} & = 
\frac{1}{N}\sum_{x,y}\alpha\frac{1-\text{SSIM}\left(I^{\mathit{left}}(x,y), 
	I^{\mathit{left}}_{\mathit{recons}}(x,y)\right)}{2} \\ & + (1-\alpha)\lVert 
I^{\mathit{left}}(x,y)-I^{\mathit{left}}_{\mathit{recons}}(x,y)\rVert_1,
\end{split}
\end{equation}
with a $3\times3$ box filter for SSIM and $\alpha$ set to $0.84$.

\textbf{Supervised loss.} 
The supervised loss measures the deviation of the predicted disparity map from the disparities estimated by Stereo DSO at a sparse set of pixels:
\begin{equation}
\mathcal{L}_s^{\mathit{left}} = \frac{1}{N}\sum_{(x,y) \in 
	\Omega_{\mathit{DSO}, \mathit{left}}}\beta_\epsilon\left(\mathit{disp}^{\mathit{left}}(x, y) - 
{\mathit{disp}}^{\mathit{left}}_{\mathit{DSO}}(x,y)\right)
\end{equation}
where $\Omega_{\mathit{DSO}, \mathit{left}}$ is the set of pixels with disparities estimated by DSO and $\beta_\epsilon(x)$ is the reverse Huber (berHu) norm introduced 
in~\cite{laina2016deeper} which lets the training focus more on larger residuals.
The threshold $\epsilon$ is adaptively set as a batch-dependent value
$\epsilon = 0.2 \max_{(x,y)\in\Omega_{\mathit{DSO}, \mathit{left}}} \left| \mathit{disp}^{\mathit{left}}(x,y) - 
\mathit{disp}_{\mathit{DSO}}^{\mathit{left}}(x,y)\right|$.

\textbf{Left-right disparity consistency loss.} Given only the left 
image as input, the network predicts the disparity map of the 
left as well as the right image as in~\cite{godard2016unsupervised}. 
As proposed in~\cite{godard2016unsupervised,zhao2015l2}, consistency between the left and right disparity image is improved by
\begin{equation}
\mathcal{L}_{\mathit{lr}}^{\mathit{left}} = \frac{1}{N}\sum_{x,y}\Bigl | \mathit{disp}^{\mathit{left}}(x, y) - 
\mathit{disp}^{\mathit{right}}(x - \mathit{disp}^{\mathit{left}}(x, y), y) \Bigr 
|.
\label{eq:lr_check}
\end{equation}

\textbf{Disparity smoothness regularization.} Depth reconstruction based on 
stereo image matching is an ill-posed problem on its own: the depth of 
homogeneously textured areas and occluded areas cannot be determined. 
For these areas, we apply the regularization term 
\begin{equation}
\mathcal{L}_{\mathit{smooth}}^{\mathit{left}}  = \frac{1}{N} \sum_{x,y}\Bigl\lvert 
\nabla^2_x \mathit{disp}^{\mathit{left}}(x, y)\Bigr\rvert e^{-\Bigl\lVert \nabla^2_x I^{\mathit{left}}(x, y) 
	\Bigr\rVert} + 
\Bigl\lvert 
\nabla^2_y \mathit{disp}^{\mathit{left}}(x, y)\Bigr\rvert e^{-\Bigl\lVert \nabla^2_y I^{\mathit{left}}(x, y) 
	\Bigr\rVert}
\end{equation}
that assumes that the predicted disparity map should be locally smooth. 
We use a second-order smoothness prior~\cite{woodford2009global} and downweight it when 
the image gradient is high~\cite{heise2013pm}.

\textbf{Occlusion regularization.} 
$\mathcal{L}_{\mathit{smooth}}^{\mathit{left}}$ itself tends to generate a 
shadow area where values gradually change from foreground to background due to 
stereo occlusion. To favor background depths and hard transitions at 
occlusions~\cite{zhong2017self}, we impose 
$\mathcal{L}_{\mathit{occ}}^{\mathit{left}}$ which penalizes the 
total sum of absolute disparities. The combination of smoothness- and 
occlusion regularizer prefers to directly take the (smaller) closeby background 
disparity which better corresponds to the assumption that the background part 
is uncovered
\begin{equation}
\mathcal{L}_{\mathit{occ}}^{\mathit{left}}  = \frac{1}{N} \sum_{x, y}\Bigl\lvert 
\mathit{disp}^{\mathit{left}}(x, y)\Bigr\rvert.
\end{equation}


\section{Deep Virtual Stereo Odometry}\label{sec:dvs_dso}

\begin{figure}[tb]
	\centering
	\includegraphics[width=\textwidth]{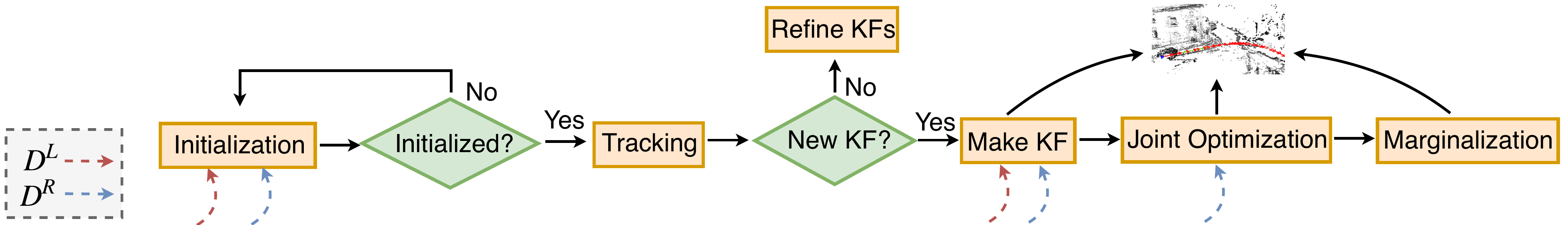}
	\caption{System overview of 
	DVSO. Every new frame is used for visual odometry and fed into 
		the proposed StackNet to predict left and right disparity. The predicted left and right disparities are used for 
		depth initialization, while the right disparity is used to form the virtual stereo term in direct sparse bundle adjustment.}
	\label{fig:sys_overview}
\end{figure}

Deep Virtual Stereo Odometry (DVSO) builds on the windowed sparse direct bundle adjustment formulation of monocular DSO.
We use our disparity predictions for DSO in two key ways:
Firstly, we initialize depth maps of new keyframes from the disparities.
Beyond this rather straight-forward approach, we also incorporate virtual direct image alignment constraints into the windowed direct bundle adjustment of DSO.
We obtain these constraints by warping images with the estimated depth by bundle adjustment and the predicted right disparities by our network assuming a virtual stereo setup. 
As shown in~\Cref{fig:sys_overview}, DVSO integrates both the predicted left disparities and right disparities for the left image. 
The right image of the stereo setup is not used for our VO method at any stage, making it a monocular VO method.

In the following, we use 
$D^{L}$ and $D^{R}$ 
as shorthands to represent the predicted left 
($disp_0^{left}$) and right disparity map 
($disp_0^{right}$) at 
scale $s = 0$, respectively.
When using purely geometric cues, scale drift is one of the main sources of 
error of monocular VO due to scale unobservability~\cite{strasdat2010scale}. 
In DVSO we use the left disparity map 
$D^{L}$ predicted by StackNet for initialization instead of randomly initializing the depth like in monocular 
DSO~\cite{engel2017direct}. 
The disparity value of an image point with coordinate $\mathbf{p}$ is converted to the inverse 
depth $d_\mathbf{p}$ using the rectified camera 
intrinsics and stereo baseline of the training set of 
StackNet~\cite{hartley2003multiple},
	$d_{\mathbf{p}} = \frac{D^{L}(\mathbf{p})}{f_xb}$.
In this way, the initialization of DVSO becomes more stable than monocular 
DSO and the depths are initialized with a consistent metric scale. 

The point selection strategy of DVSO is similar to 
monocular DSO~\cite{engel2017direct}, while we also introduce a left-right 
consistency check (similar to \Cref{eq:lr_check}) to filter out the 
pixels which likely lie in the occluded area
\begin{equation}
	e_{lr} = \Bigl | D^{L}(\mathbf{p}) - 
	D^{R}(\mathbf{p'}) \Bigr | \quad \text{with} \quad \mathbf{p'} = \mathbf{p} 
	- 
	\begin{bmatrix}
	D^L(\mathbf{p}) & 0\\
	\end{bmatrix}^\top.
\end{equation}
The pixels with $e_{lr} > 1$ are not selected. 

Every new frame is firstly tracked with respect to the 
reference keyframe using direct image alignment in a coarse-to-fine 
manner~\cite{engel2017direct}. Afterwards  DVSO 
decides if a new keyframe has to be created for the new frame following the criteria proposed 
by~\cite{engel2017direct}. When a new keyframe is created, 
the temporal multi-view energy function
		$E_{photo} := \sum_{i\in\mathcal{F}}\sum_{\mathbf{p} \in 
		\mathcal{P}_i}\sum_{j \in \text{obs}(\mathbf{p})}E_{ij}^{\mathbf{p}}$
needs to be optimized, where $\mathcal{F}$ is a fixed-sized window containing 
the active keyframes, $\mathcal{P}_i$ is 
the set of points selected from its host keyframe with index $i$ and $j \in 
\text{obs}(\mathbf{p})$ is the index of the keyframe which observes 
$\mathbf{p}$. $E_{ij}^{\mathbf{p}}$ is the photometric error of 
the point $\mathbf{p}$ when projected from the host keyframe $I_i$ onto the 
other keyframe $I_j$:
\begin{equation}
	E_{ij}^{\mathbf{p}} := \omega_{\mathbf{p}} 
		\left\| (I_j[\mathbf{\tilde{p}}] - 
	b_j) - \frac{e^{a_j}}{e^{a_i}}(I_i[\mathbf{p}] - b_i) \right\|_\gamma,
	\label{eq:5.1}
\end{equation} 
where $\mathbf{\tilde{p}}$ is the projected image coordinate using the relative 
rotation 
matrix $\textbf{R} \in SO(3)$ and translation vector $\textbf{t} \in 
\mathbb{R}^3$~\cite{hartley2003multiple},
$\mathbf{\tilde{p}} = \Pi_c\left(\mathbf{R}\Pi_c^{-1}\left(\mathbf{p}, d_{\mathbf{p}}\right) 
+ \mathbf{t}\right)$,
where $\Pi_\mathbf{c}$ and  $\Pi_\mathbf{c}^{-1}$ are the camera projection and 
back-projection functions. The parameters $a_i$, $a_j$, $b_i$ and 
$b_j$ are used for modeling the affine brightness 
transformation~\cite{engel2017direct}. The weight 
$\omega_{\mathbf{p}}$ penalizes the points with high image 
gradient~\cite{engel2017direct} with the intuition that the error originating 
from bilinear interpolation of the discrete image values is larger. $\lVert 
\cdot \rVert_{\gamma}$ is the Huber norm with the threshold $\gamma$. For the 
detailed explanation of the energy function, please refer 
to~\cite{engel2017direct}.

To further improve the accuracy of DVSO, inspired by Stereo 
DSO~\cite{wang2017stereoDSO} which couples the static stereo 
term with the temporal multi-view energy function, we introduce a novel 
\emph{virtual stereo} term $E^{\dagger \mathbf{p}}$ for each point $\mathbf{p}$
\begin{equation}
		E_{i}^{\dagger\mathbf{p}} = \omega_{\mathbf{p}} 
	\left\| I_i^\dagger\left[\mathbf{p}^\dagger\right] - I_i\left[\mathbf{p}\right] \right\|_\gamma 
	\quad 
	\text{with} \quad I_i^\dagger\left[\mathbf{p}^\dagger\right] = I_i\left[\mathbf{p}^\dagger 
	- 
	\begin{bmatrix}
	D^{R}\left(\mathbf{p}^\dagger\right) & 0 \\
	\end{bmatrix}^\top\right],
	\label{eq:5.8}
\end{equation}
where $\mathbf{p}^\dagger = \Pi_\mathbf{c}(\Pi_\mathbf{c}^{-1}(\mathbf{p}, 
d_{\mathbf{p}}) + \mathbf{t}_b)$ is the virtual projected coordinate of 
$\mathbf{p}$ using the  
vector $\mathbf{t}_b$ denoting the virtual stereo baseline which is known 
during the training of StackNet. 
The intuition behind this term is to optimize the estimated depth of the visual odometry to become consistent with the disparity prediction of StackNet.
Instead of imposing the consistency directly on the estimated and predicted disparities, we formulate the residuals in photoconsistency which better reflects the uncertainties of the prediction of StackNet and also keeps the unit of the residuals consistent with the temporal direct image alignment terms.

We then optimize the total energy 
\begin{equation}
E_{photo} := \sum_{i\in\mathcal{F}}\sum_{\mathbf{p} \in 
	\mathcal{P}_i} \left( \lambda E_{i}^{\dagger\mathbf{p}} + \sum_{j \in 
	\text{obs}(\mathbf{p})}E_{ij}^{\mathbf{p}} \right) \ ,
\label{eq:total_func}
\end{equation}
where the coupling factor $\lambda$ balances the temporal and the 
virtual stereo term. All the parameters of the total energy are 
jointly optimized using the Gauss Newton method~\cite{engel2017direct}. In order to 
keep a fixed size of the active window ($N=7$ keyframes in our experiments), 
old keyframes are removed from the 
system by marginalization using the Schur complement~\cite{engel2017direct}. 
Unlike sliding window bundle adjustment, the parameter estimates 
outside the optimization window including camera poses and depths in a 
marginalization prior are also incorporated into the optimization. In contrast 
to the MSCKF~\cite{msckf}, the depths of pixels are explicitly maintained in 
the state and 
optimized for. In our optimization framework we trade off predicted depth and 
triangulated depth using robust norms.


\section{Experiments}\label{sec:exp}
We quantitatively evaluate our StackNet with 
other state-of-the-art monocular depth 
prediction methods on the 
publicly available KITTI 
dataset~\cite{Geiger2013IJRR}. In the supplementary materials, we 
demonstrate results on the Cityscapes 
dataset~\cite{Cordts2016Cityscapes} and the 
Make3D dataset~\cite{saxena2006learning} to 
show the generalization ability. 
For DVSO, we evaluate its tracking
accuracy on the KITTI odometry benchmark with 
other state-of-the-art monocular as well as 
stereo visual odometry systems. In the supplementary material, we also 
demonstrate its results on the \textit{Frankfurt} sequence of the Cityscapes 
dataset to show the generalization of DVSO.

\begin{figure}[tb]
	\centering
	\begin{tabular}
		{ccccccc}
		\scalebox{.6} 
		{Input}&\scalebox{.6}{GT}&\scalebox{.6} {Ours}& 
		\scalebox{.6}{Kuznietsov 
		et 
		al.\cite{kuznietsov2017semi}}&\scalebox{.6}{Godard et al.
		\cite{godard2016unsupervised}}
		&\scalebox{.6}{Garg et al.\cite{garg2016unsupervised}}&\scalebox{.6} 
		{Eigen et al.\cite{eigen2014depth}}\\
		\includegraphics[width=0.135\textwidth]{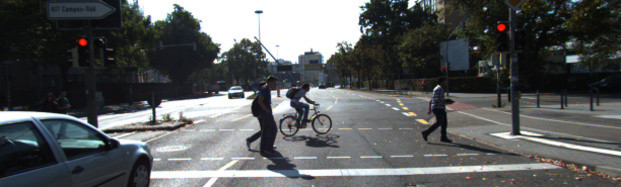}&
		\includegraphics[width=0.135\textwidth]{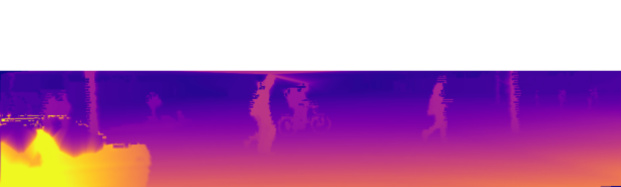} &
		\includegraphics[width=0.135\textwidth]{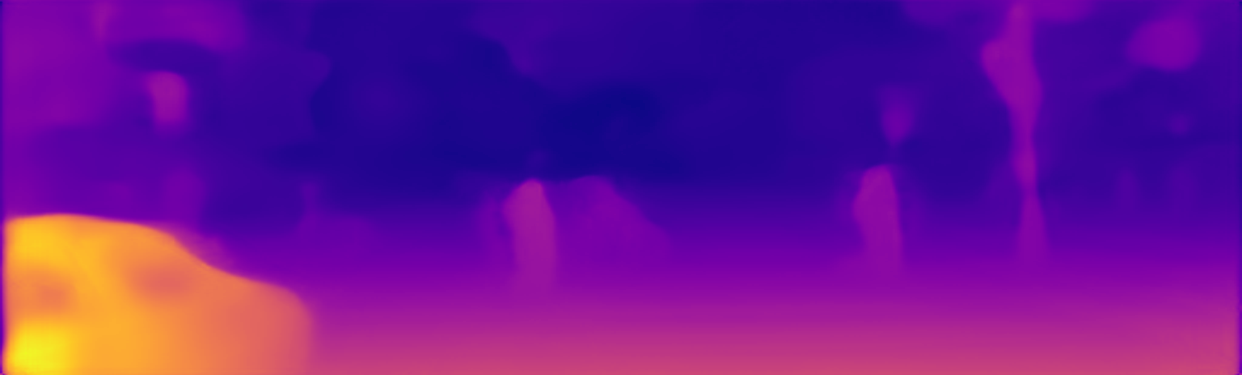} &
		\includegraphics[width=0.135\textwidth]{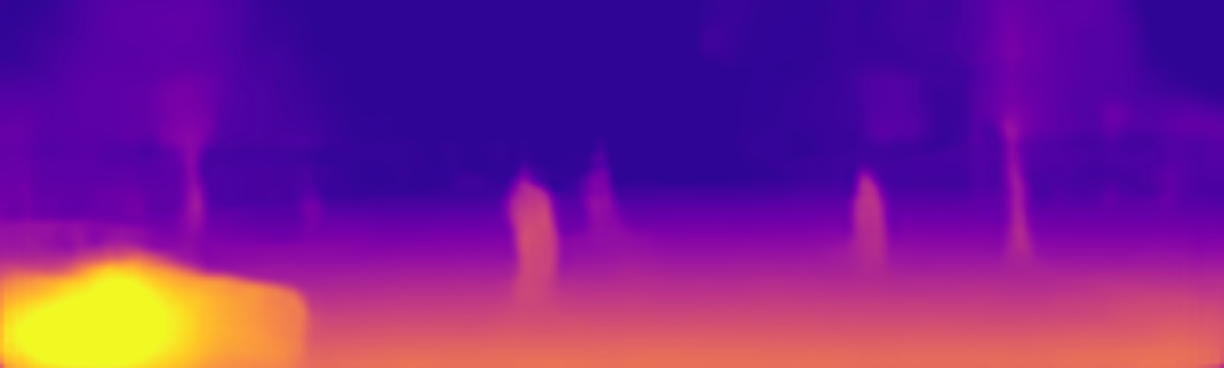} &
		\includegraphics[width=0.135\textwidth]{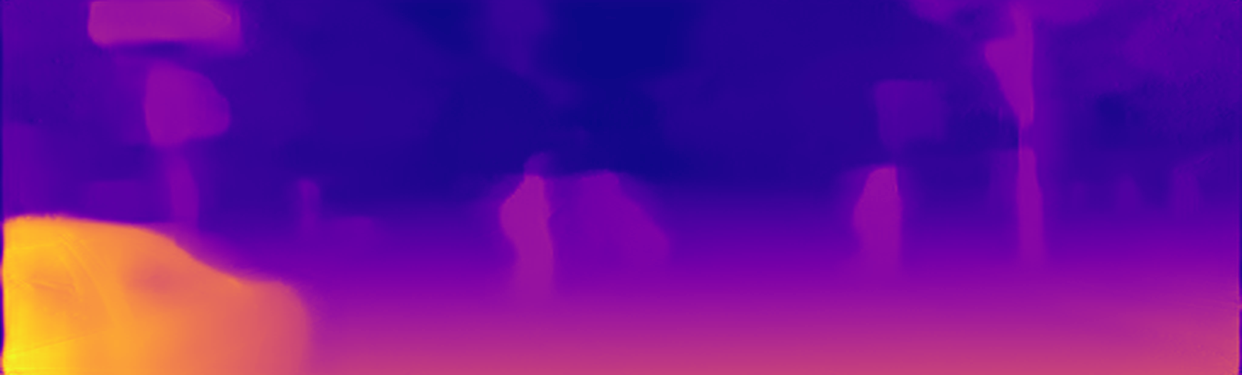}
		 &
		\includegraphics[width=0.135\textwidth]{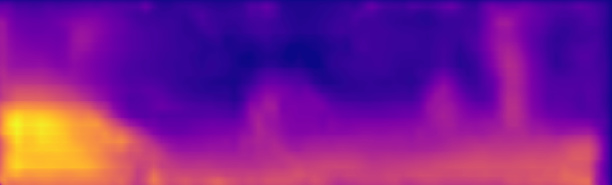} &
		\includegraphics[width=0.135\textwidth]{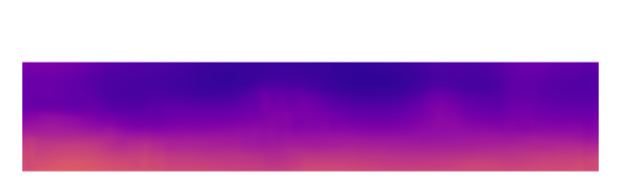}\\
		\includegraphics[width=0.135\textwidth]{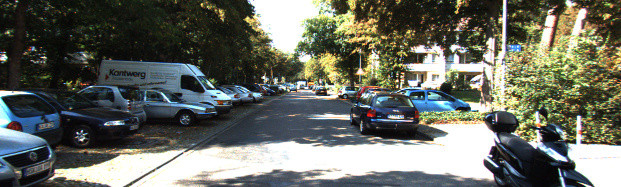}&
		\includegraphics[width=0.135\textwidth]{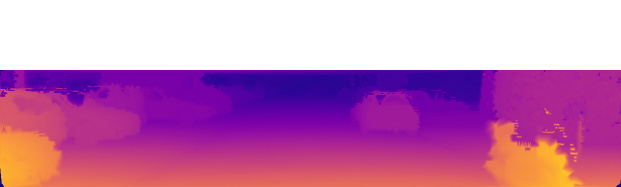}&
		\includegraphics[width=0.135\textwidth]{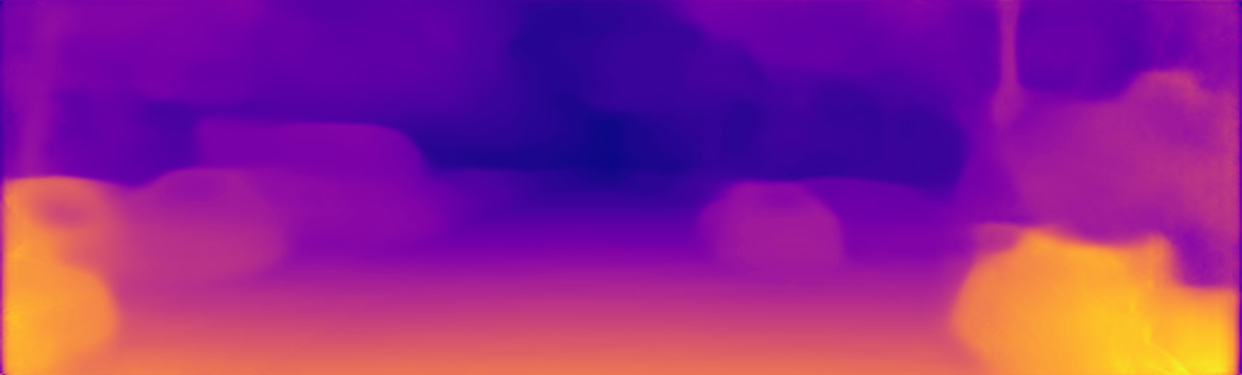}&
		\includegraphics[width=0.135\textwidth]{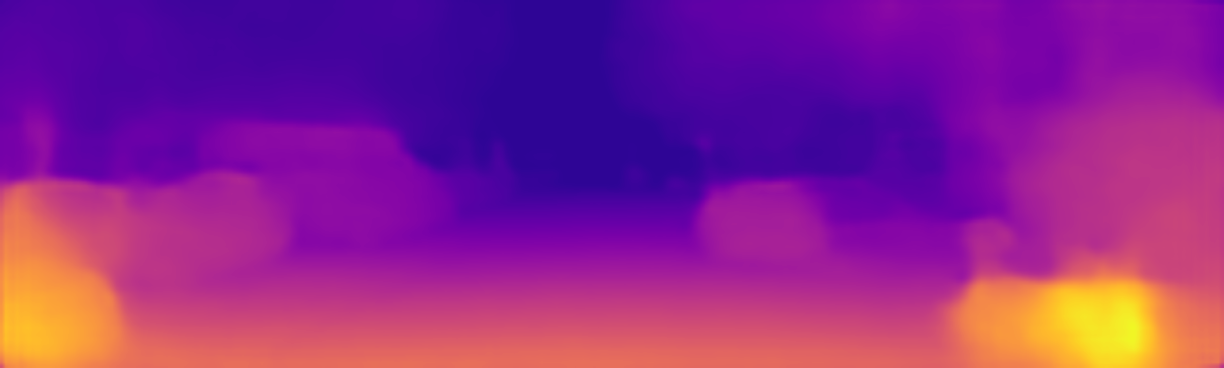}&
		\includegraphics[width=0.135\textwidth]{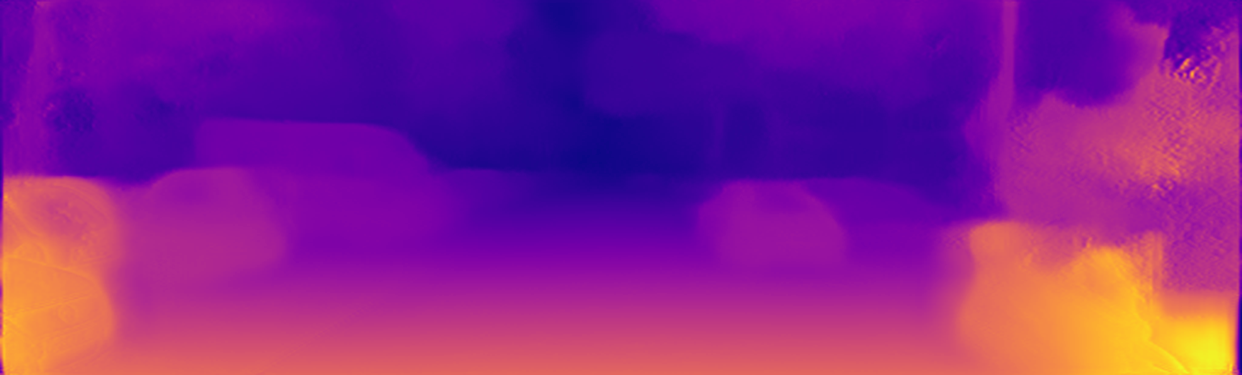}&
		\includegraphics[width=0.135\textwidth]{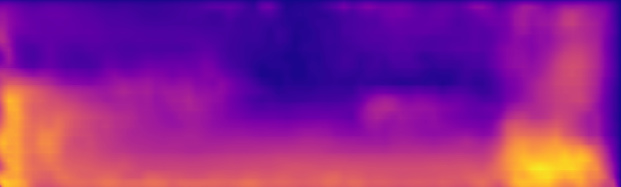}&
		\includegraphics[width=0.135\textwidth]{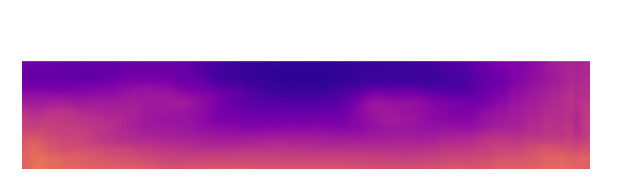}\\
		\includegraphics[width=0.135\textwidth]{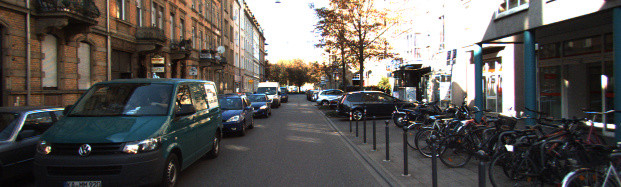}&
		\includegraphics[width=0.135\textwidth]{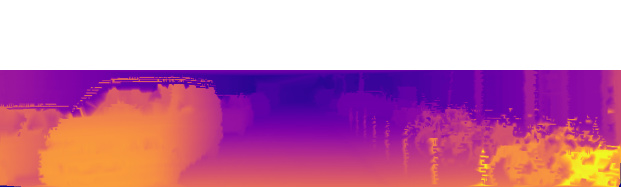}&
		\includegraphics[width=0.135\textwidth]{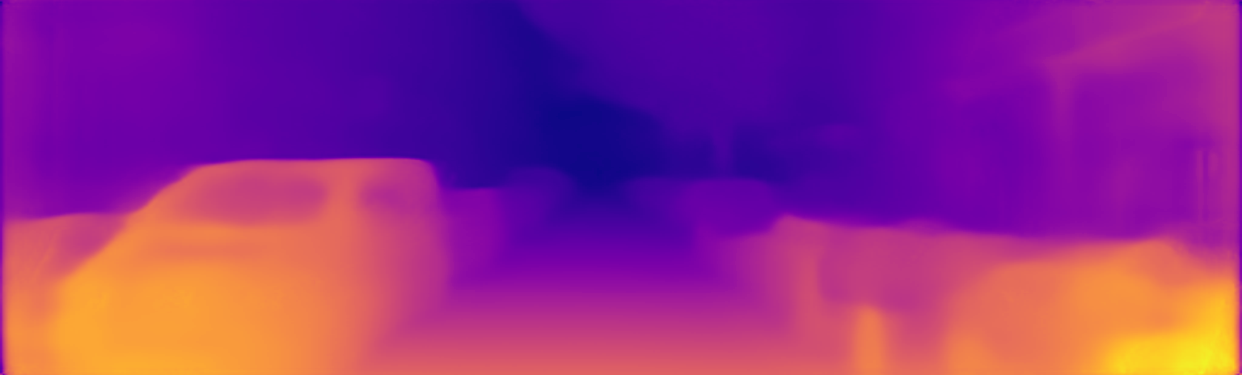}&
		\includegraphics[width=0.135\textwidth]{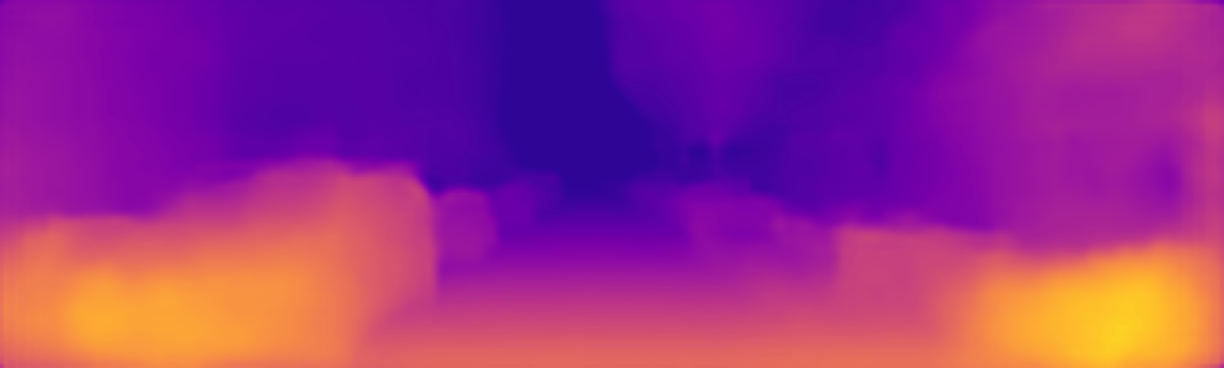}&
		\includegraphics[width=0.135\textwidth]{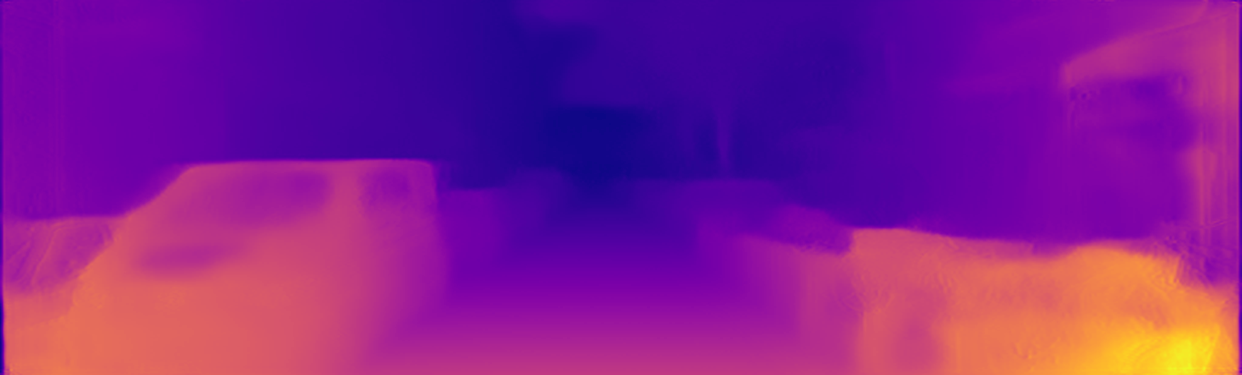}&
		\includegraphics[width=0.135\textwidth]{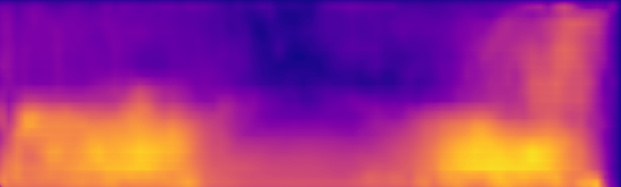}&
		\includegraphics[width=0.135\textwidth]{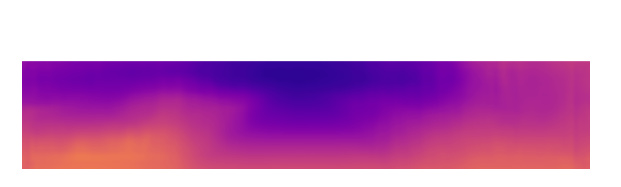}\\
		
	\end{tabular}
\caption{Qualitative comparison with 
	state-of-the-art methods. The ground  
	truth is interpolated for better visualization. Our approach shows  
	better  
	prediction on thin structures than the self-supervised  
	approach~\cite{godard2016unsupervised}, and delivers more detailed 
	disparity maps 
	than the semi-supervised approach using LiDAR 
	data~\cite{kuznietsov2017semi}.}
\label{fig:compare_all_few}
\end{figure}

\subsection{Monocular Depth Estimation}
\textbf{Dataset.} 
We train StackNet using the train/test split (\textbf{K}) of 
Eigen et al.~\cite{eigen2014depth}. 
The training set contains 23488 images from 
28 scenes belonging to the categories "city", 
"residential" and "road". We used 22600 images of 
them for training and the remaining ones for validation. We 
further split \textbf{K} into 2 subsets $\text{\textbf{K}}_o$ and 
$\text{\textbf{K}}_r$. 
$\text{\textbf{K}}_o$ contains the images of the sequences 
which appear in the training set (but not the test set) of 
the KITTI odometry benchmark on which we use 
Stereo DSO~\cite{wang2017stereoDSO} to extract 
sparse ground-truth depth data. 
$\text{\textbf{K}}_r$ 
contains the remaining images in 
\textbf{K}. 
Specifically, $\text{\textbf{K}}_o$ contains the images of 
sequences 01, 02, 06, 08, 09 and 10 of the KITTI 
odometry benchmark.


\textbf{Implementation details.} 
StackNet is implemented in 
TensorFlow~\cite{abadi2016tensorflow} and 
trained from scratch on a single Titan X 
Pascal GPU. 
We resize the images to $512 \times 
256$ for training and it takes less than 40ms 
for inference including the I/O overhead. The 
weights are set to $\alpha_u = 1$, $\alpha_s 
= 10$, $\alpha_{lr} = 1$, $\alpha_{smooth} = 
0.1/2^s$ and $\alpha_{occ} = 0.01$, where $s$ 
is the output scale. As suggested 
by~\cite{godard2016unsupervised}, we 
use exponential linear units (ELUs) 
for SimpleNet, while we use leaky 
rectified linear units (Leaky ReLUs) for ResidualNet. We 
first train SimpleNet on $\text{\textbf{K}}_o$ in 
the semi-supervised way for 80 epochs 
with a batch size of 8 using the Adam 
optimizer~\cite{kingma2014adam}. The 
learning rate is initially set to 
$\lambda = 10^{-4}$ for the first 50 epochs 
and halved every 15 epochs 
afterwards until the end. Then we train SimpleNet with $\lambda = 5 \times 
10^{-5}$ on $\text{\textbf{K}}_r$ for 40 
epochs in the self-supervised way without 
$\mathcal{L_S}$. In the end, we 
train again on 
$\text{\textbf{K}}_o$ without $\mathcal{L_U}$ using $\lambda = 10^{-5}$ for 5 
epochs. We explain the dataset schedule as well as the parameter 
tuning in detail in the 
\href{https://vision.in.tum.de/_media/spezial/bib/yang2018dvso-supp.pdf}{supplementary
	material}.

After training SimpleNet, we freeze 
its weights and train StackNet by cascading 
ResidualNet. StackNet is trained with 
$\lambda = 5 \times 
10^{-5}$ in the same dataset 
schedules but with less epochs, i.e. 30, 15, 3 
epochs, respectively. We apply random gamma, brightness and color 
augmentations~\cite{godard2016unsupervised}. We 
also employ the post-processing for left disparities proposed by Godard et 
al.~\cite{godard2016unsupervised} to reduce the effect of stereo disocclusions.
In the 
\href{https://vision.in.tum.de/_media/spezial/bib/yang2018dvso-supp.pdf}{supplementary
	material} we also provide an ablation study on the various loss terms.

\textbf{KITTI.}~\Cref{tab:6.1} 
shows the evaluation results with the error 
metrics used in~\cite{eigen2014depth}. We 
crop the images as applied by Eigen et al.~\cite{eigen2014depth} to compare 
with \cite{godard2016unsupervised}, \cite{kuznietsov2017semi} within different 
depth ranges.
The best performance of our network is achieved 
with the dataset schedule 
$\text{\textbf{K}}_o 
\rightarrow 
\text{\textbf{K}}_r \rightarrow 
\text{\textbf{K}}_o $ as we described above. 
We outperform the state-of-the-art 
self-supervised approach 
proposed by Godard et 
al.~\cite{godard2016unsupervised} by a large 
margin.
Our method also 
outperforms the state-of-the-art semi-supervised method using 
the LiDAR ground truth proposed by 
Kuznietsov et al.~\cite{kuznietsov2017semi} 
on all the metrics except for the less restrictive $\delta < 
1.25^2$ and $\delta < 1.25^3$. 

\begin{figure}[tb]
	\centering
	\begin{subfigure}[t]{.24\textwidth}
		\centering
		\caption*{\tiny Input}
		\includegraphics[width=\textwidth]{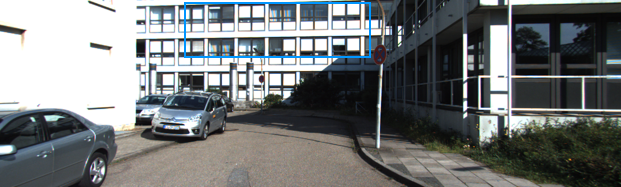}
	\end{subfigure}
	\begin{subfigure}[t]{.24\textwidth}
		\centering
		\caption*{\tiny Ours}
		\includegraphics[width=\textwidth]{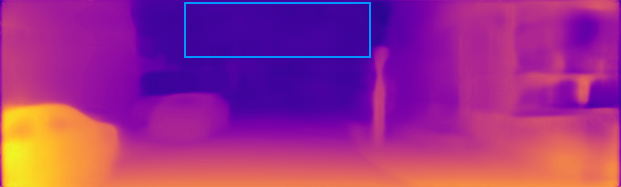}
	\end{subfigure}
	\begin{subfigure}[t]{.24\textwidth}
		\centering
		\caption*{\tiny Godard et al.~\cite{godard2016unsupervised}}
		\includegraphics[width=\textwidth]{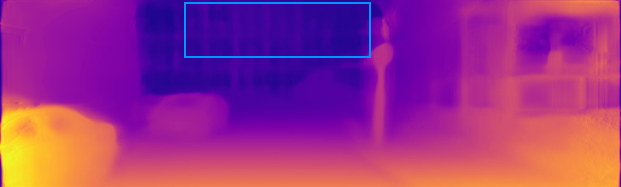}
	\end{subfigure}
	\begin{subfigure}[t]{.24\textwidth}
		\centering
		\caption*{\tiny Kuznietsov et al.~\cite{kuznietsov2017semi}}
		\includegraphics[width=\textwidth]{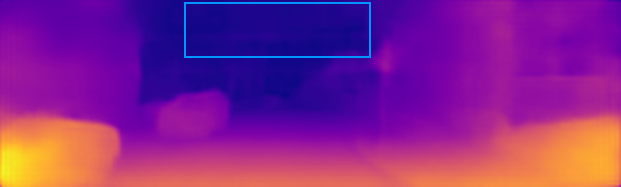}
	\end{subfigure}
	\begin{subfigure}[t]{.24\textwidth}
		\centering
		\includegraphics[width=\textwidth]{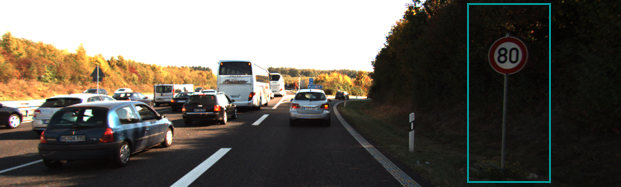}
	\end{subfigure}
	\begin{subfigure}[t]{.24\textwidth}
		\centering
		\includegraphics[width=\textwidth]{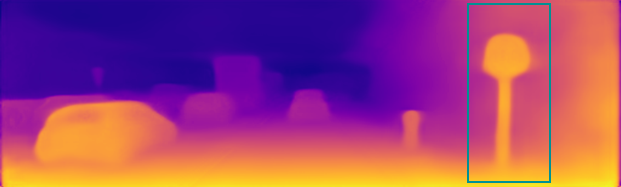}
	\end{subfigure}
	\begin{subfigure}[t]{.24\textwidth}
		\centering
		\includegraphics[width=\textwidth]{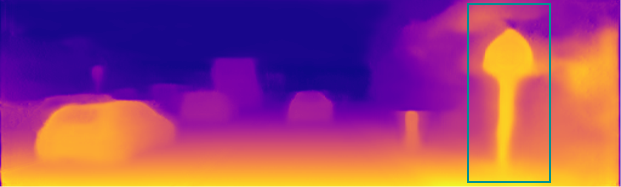}
	\end{subfigure}
	\begin{subfigure}[t]{.24\textwidth}
		\centering
		\includegraphics[width=\textwidth]{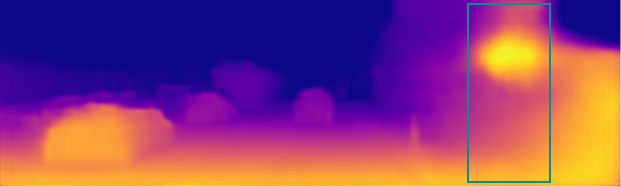}
	\end{subfigure}
	\caption{Qualitative results on Eigen et al.'s KITTI Raw test split. The result of Godard et 
		al.~\cite{godard2016unsupervised} shows a strong shadow effect around 
		object 
		contours, while our result does not. The result of 
		Kuznietsov~\cite{kuznietsov2017semi} shows failure on predicting the 
		traffic sign. Both other methods ~\cite{godard2016unsupervised,kuznietsov2017semi} 
		predict 
		checkerboard artifacts on the far 
		building, while our approach predicts such artifacts less.}
	\label{fig:shadow}
\end{figure}

\Cref{fig:compare_all_few}
shows a qualitative comparison with other 
state-of-the-art methods. Compared to the 
semi-supervised approach, 
our results contain more details and deliver
comparable prediction on thin structures like the poles.
Although the results of Godard et 
al.~\cite{godard2016unsupervised} 
appear more detailed on some parts, they are 
not actually accurate, which can be inferred by the quantitative 
evaluation. In general, the predictions of Godard et 
al.~\cite{godard2016unsupervised} on 
thin objects are not as accurate as our method. In the 
\href{https://vision.in.tum.de/_media/spezial/bib/yang2018dvso-supp.pdf}{supplementary
	material}, we show the error
maps for the predicted depth maps.
~\Cref{fig:shadow} further show 
the advantages of our method compared to the 
state-of-the-art self-supervised and semi-supervised
approaches. The results of Godard et 
al.~\cite{godard2016unsupervised} are 
predicted by the network trained with both 
the Cityscapes dataset and the KITTI dataset.
On the wall of the far building 
in the left figure, our network can 
better predict consistent depth on the 
surface, while the prediction of the 
self-supervised network shows strong checkerboard 
artifact, which is apparently inaccurate. The 
semi-supervised approach also shows checkerboard artifact (but much slighter). 
The right side of the figure shows shadow artifacts for the approach of Godard et al.~\cite{godard2016unsupervised} around the boundaries of the traffic 
sign, while the result of Kuznietsov et al.~\cite{kuznietsov2017semi} fails to predict the structure. 
Please refer to our 
\href{https://vision.in.tum.de/_media/spezial/bib/yang2018dvso-supp.pdf}{supplementary
	material} for further results.
We also demonstrate how our trained depth prediction network generalizes to 
other datasets in the 
\href{https://vision.in.tum.de/_media/spezial/bib/yang2018dvso-supp.pdf}{supplementary
	material}.

\begin{table}[tb]
	\centering
	\tiny
	\begin{tabular}{lcccccccccc}
		\hline
		&&&&&&&&&&\\
		&    & &RMSE   & RMSE (log)   & ARD  & SRD  &  & $\delta < 1.25$ & 
		$\delta < 1.25^2$ & $\delta < 1.25^3$ \\ \cline{4-7} \cline{9-11} 
		Approach & Dataset &  & \multicolumn{4}{c}{lower is better} &  & 
		\multicolumn{3}{c}{higher is better} \\
		\hline
		Godard et al.~\cite{godard2016unsupervised}, ResNet  & 
		\textbf{CS}$\rightarrow$\textbf{K}&& 
		4.935 & 0.206 & 0.114 & 0.898 && 
		0.861 & 0.949 & 0.976 \\
		Kuznietsov et al.~\cite{kuznietsov2017semi} &\textbf{K}&& 
		\textit{4.621} & \textit{0.189} & 
		0.113 & \textit{0.741} && 
		0.862 & \textbf{0.960} & \textbf{0.986} \\
		Ours, SimpleNet  &$\text{\textbf{K}}_o$&& 4.886 & 0.209 & 
		0.112 & 
		0.888 && 
		0.862 & 0.950 & 0.976 \\
		Ours, SimpleNet 
		&$\text{\textbf{K}}_o\rightarrow\text{\textbf{K}}_r$&& 4.817 & 
		0.202 & 
		0.108 
		& 0.862 && 
		\textit{0.867} & 0.950 & 0.977 \\
		Ours, SimpleNet 
		&$\text{\textbf{K}}_r\rightarrow\text{\textbf{K}}_o$&& 4.890 & 
		0.208 & 
		0.115 
		& 0.870 && 
		0.863 & 0.950 & 0.977 \\
		Ours, SimpleNet  
		&$\text{\textbf{K}}_o\rightarrow\text{\textbf{K}}_r\rightarrow\text{\textbf{K}}_o$&&
		
		4.785 & 
		0.199 & \textit{0.107} & 
		0.852 && 
		0.866 & 0.950 & 0.978 \\
		Ours, StackNet 
		&$\text{\textbf{K}}_o\rightarrow\text{\textbf{K}}_r\rightarrow\text{\textbf{K}}_o$&&
		
		\textbf{4.442} & \textbf{0.187} & 
		\textbf{0.097} & \textbf{0.734} && 
		\textbf{0.888} & \textit{0.958} & \textit{0.980} \\
		\hline
		\hline
		Garg et al.~\cite{garg2016unsupervised} L12 Aug 8$\times$
		&\textbf{K}&& 
		5.104 & 0.273 & 0.169 & 1.080 
		&& 0.740 & 0.904 & 0.962 \\
		Godard et al.~\cite{godard2016unsupervised}, ResNet 
		&\textbf{CS}$\rightarrow$\textbf{K}&& 
		3.729 & 0.194 & 0.108 & 
		0.657 && 0.873 & 0.954 & 0.979 \\
		Kuznietsov et al.~\cite{kuznietsov2017semi} &\textbf{K}&& 
		\textit{3.518} & 
		\textit{0.179} & 
		\textit{0.108} & \textit{0.595} 
		&& \textit{0.875} & \textbf{0.964} & \textbf{0.988} \\
		Ours, StackNet
		&$\text{\textbf{K}}_o\rightarrow\text{\textbf{K}}_r\rightarrow\text{\textbf{K}}_o$&&
		
		\textbf{3.390} & \textbf{0.177} & 
		\textbf{0.092} & \textbf{0.547} && 
		\textbf{0.898} & \textit{0.962} & \textit{0.982} \\
		\hline      
	\end{tabular}
	\caption{
	Evaluation results on the KITTI~\cite{Geiger2012CVPR} Raw test split 
		of Eigen et 
		al.~\cite{eigen2014depth}. \textbf{CS} refers to the Cityscapes 
		dataset~\cite{Cordts2016Cityscapes}. Upper part: depth range 0-80\,m, 
		lower part: 1-50\,m.
All results are obtained using the crop 
		from~\cite{eigen2014depth}. Our SimpleNet trained on 
		$\text{\textbf{K}}_o$ outperforms~\cite{godard2016unsupervised} 
		(self-supervised) trained on 
		\textbf{CS} and 
		\textbf{K}. StackNet also outperforms 
		semi-supervision with LiDAR~\cite{kuznietsov2017semi} on most metrics.}
	\label{tab:6.1}
\end{table}

\subsection{Monocular Visual Odometry}
\textbf{KITTI odometry benchmark}. 
The KITTI odometry benchmark contains 11 (0-10)
training sequences and 11 (11-21) test sequences. Ground-truth 6D poses are 
provided for the training sequences, whereas for the test sequences evaluation results
are obtained by submitting to the KITTI website. 
We use the error metrics proposed in~\cite{Geiger2013IJRR}.

We firstly provide an ablation study for DVSO to show the effectiveness 
of the design choices in our approach. 
In~\Cref{tab:diff_settings} we give results for DVSO in different variants with the following components:
initializing the depth with the left disparity prediction ($in$), using the right disparity for the virtual 
stereo term in windowed bundle adjustment ($vs$), checking left-right disparity consistency for point selection ($lr$),
and tuning the virtual stereo baseline $tb$.
The intuition behind the virtual stereo baseline is that StackNet is trained over various 
camera parameters and hence provides a depth scale for an average baseline.
For $tb$, we therefore tune the scale factors of different sequences with different cameras intrinsics, to better align the estimated scale with the
ground truth.
Baselines are tuned for each of the 3 different camera 
parameter sets 
in the training set individually using grid search on one training sequence. 
Specifically, we tuned the baselines on sequences 00, 03 and 05 which 
correspond to 3 different camera parameter sets. The test set 
contains the same camera parameter sets as the training 
set and we map the virtual baselines for $tb$ correspondingly. Monocular DSO 
(after Sim(3) alignment) is also 
shown as the baseline. 
The results show that our full approach achieves the 
best average performance.
Our StackNet also adds significantly to the performance of DVSO compared with 
using depth predictions from~\cite{godard2016unsupervised}.

\begin{table}[tb]
	\centering
	\tiny
	\begin{tabular}{c|cc|cc|cc|cc|cc|cc|cc}
		& \multicolumn{2}{c|}{Mono DSO} & 
		\multicolumn{2}{c|}{$in$} & 
		\multicolumn{2}{c|}{$in,vs$} 
		& 
		\multicolumn{2}{c|}{$in,vs,lr$} & 
		\multicolumn{2}{c|}{$in,vs,tb$}&
		\multicolumn{2}{c|}{DVSO'(\cite{godard2016unsupervised})}&
		\multicolumn{2}{c}{DVSO}\\ 
		\hline
		Seq. & $t_{rel}$ & $r_{rel}$ & $t_{rel}$ & $r_{rel}$ & $t_{rel}$ & 
		$r_{rel}$ & $t_{rel}$ & $r_{rel}$& $t_{rel}$ & $r_{rel}$& $t_{rel}$ & 
		$r_{rel}$& $t_{rel}$ & $r_{rel}$\\
		00$^\dagger$   & 188 & \textit{0.25} & 13.1 & 0.30 & 0.95 & 
		\textbf{0.24} & 0.93 & \textbf{0.24}& 
		\textit{0.73} & \textit{0.25}& 1.02 & 0.28& \textbf{0.71} & 
		\textbf{0.24}
		\\
		03$^\dagger$    & 17.7 & \textbf{0.17} & 9.10 & 0.29 & 2.56 & 0.19 
		&2.56 & \textit{0.18}& \textbf{0.78} & 0.19& 4.78 & 0.18& \textit{0.79} 
		& 
		\textit{0.18}
		\\
		04$^\dagger$    & 0.82 & 0.16 & 0.83 & 0.29 & 0.69 & \textbf{0.06} 
		&0.67 & \textit{0.07}& 
		\textit{0.36} & \textbf{0.06}& 2.03 & 0.14& \textbf{0.35} & 
		\textbf{0.06}
		\\
		05$^\dagger$    & 72.6 & \textit{0.23} & 12.7 & \textit{0.23} & 0.67 & 
		\textit{0.23} &0.64 & \textit{0.23}& 
		\textit{0.61} & 0.23& 2.11 & \textbf{0.21}& \textbf{0.58} & 0.22
		\\
		07$^\dagger$    & 48.4 & \textbf{0.32} & 18.5 & 0.91 & 0.85 & 0.41 & 
		\textit{0.80} & 0.38&0.81 & 0.40& 1.09 & 0.39& \textbf{0.73} & 
		\textit{0.35}
		\\
		\hline
		01$^\ast$   & 9.17 & \textit{0.12} & 4.30 & 0.41 & 1.50 & \textbf{0.11} 
		&1.52 & 
		\textit{0.12}& 
		\textbf{1.15} 
		& \textbf{0.11}& 1.23 & \textbf{0.11}& \textit{1.18} & \textbf{0.11}
		\\
		02$^\ast$   & 114 & \textbf{0.22} & 9.58 & 0.26 & 1.08 & \textit{0.23} 
		&1.05 & 
		\textit{0.23}& \textit{0.86} & \textit{0.23}& 0.87 & \textit{0.23}& 
		\textbf{0.84} & 
		\textbf{0.22}
		\\
		06$^\ast$   & 42.2 & \textbf{0.20} & 11.2 & 0.30 & 0.84 & \textit{0.23} 
		&0.80 & 
		0.24& 
		\textit{0.73} & \textit{0.23}& 0.87 & 0.24& 
		\textbf{0.71} & 
		\textbf{0.20}
		\\
		08$^\ast$   & 177 & \textit{0.26} & 14.9 & 0.28 & 1.11 & \textit{0.26} 
		& 1.10 & 
		\textit{0.26}&\textit{1.05} 
		& 
		\textit{0.26}& \textit{1.05} & \textit{0.26}& \textbf{1.03} & 
		\textbf{0.25}
		\\
		09$^\ast$   & 28.1 & \textbf{0.21} & 14.2 & \textit{0.23} & 1.03 & 
		\textbf{0.21} & 0.95 & 
		\textbf{0.21}&0.88
		& 
		\textbf{0.2}1& \textit{0.87} & \textbf{0.21}& \textbf{0.83} & 
		\textbf{0.21}
		\\
		10$^\ast$   & 24.0 & \textit{0.22} & 9.93 & 0.27 & \textbf{0.58} & 0.23 
		&\textit{0.59} & 
		\textit{0.22}& 0.74 & \textit{0.22}& \textbf{0.68} & \textbf{0.21}& 
		0.74 & \textbf{0.21}
		\\ 
		\hline
		\hline
		mean$^\dagger$ & 65.50 & 0.23 & 10.85 & 0.40 & 1.14 & 0.23& 1.12 & 
		\textit{0.22}&\textit{0.66} & 
		0.23& 2.21 & 0.24& 
		\textbf{0.63} & 
		\textbf{0.21}
		\\
		mean$^\ast$ & 65.75 & \textit{0.21} & 10.69 & 0.29 & 1.02 & 
		\textit{0.21}& 1.00 & 
		\textit{0.21}&\textit{0.90} & 
		\textit{0.21}&0.93 & \textit{0.21}& 
		\textbf{0.89} & 
		\textbf{0.20}
		\\
		overall mean & 65.64 & \textit{0.21} & 10.76 & 0.34 & 1.08 & 0.22& 1.06 
		& 
		0.22&\textit{0.79} & 
		0.22& 1.51 & 0.22& 
		\textbf{0.77} & 
		\textbf{0.20}
		\\
	\end{tabular}
	\caption{Ablation study for DVSO. $^\ast$ and $^\dagger$ indicate the 
	sequences 
	used and not used for training StackNet, respectively. $t_{rel}$($\%$) and 
		$r_{rel}$($^\circ$) are 
		translational-
		 and rotational RMSE, respectively. Both $t_{rel}$ and 
		$r_{rel}$
		are
		averaged over 100 to 800\,m intervals. $in$: $D^{L}$ is used for 
		depth initialization. $vs$: virtual stereo term is used with $D^{R}$. 
		$lr$: left-right disparity consistency is checked using predictions. 
		$tb$: tuned virtual baseline is used. 
		DVSO'(\cite{godard2016unsupervised}): full ($in,vs,lr,tb$) with 
		depth 
		from~\cite{godard2016unsupervised}. DVSO: full with 
		depth from StackNet. Best results are shown as bold, 
		second 
		best italic. DVSO clearly outperforms the other variants.}
	\label{tab:diff_settings}
\end{table}

\begin{SCfigure}
	\centering
		\includegraphics[width=.60\textwidth]{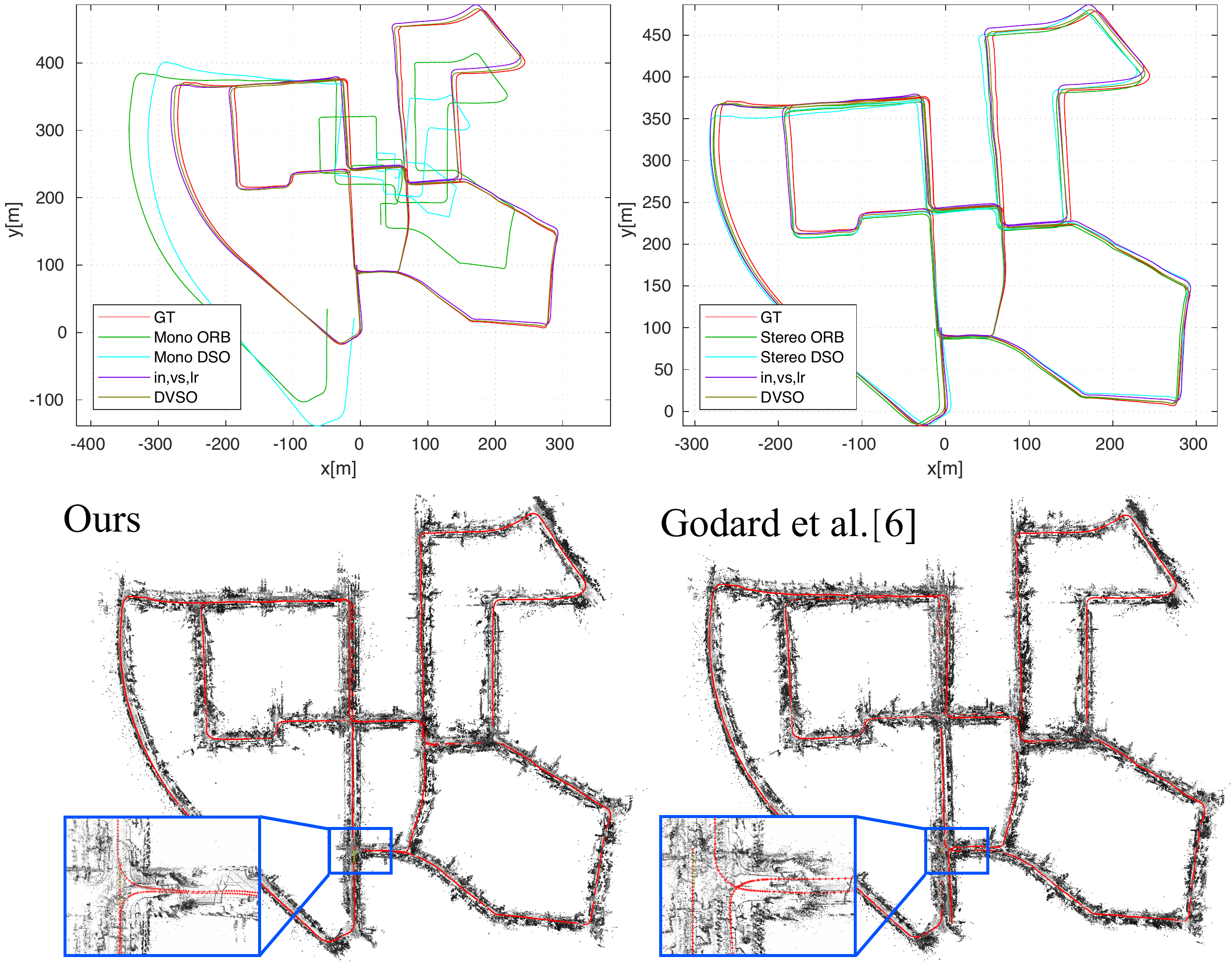}
	\caption{Results on KITTI odometry seq. 00. Top: 
		comparisons with monocular methods (Sim(3)-aligned) and stereo methods. 
		DVSO provides significantly more consistent 
		trajectories than other monocular methods and compares well to 
		stereo approaches. Bottom: DVSO with StackNet produces more accurate 
		trajectory and map than with~\cite{godard2016unsupervised}. }
	\label{fig:kitti_traj}
\end{SCfigure}

\begin{figure}[tb]
	\centering
	\begin{subfigure}[t]{.49\textwidth}
		\centering
		\includegraphics[width=\textwidth]{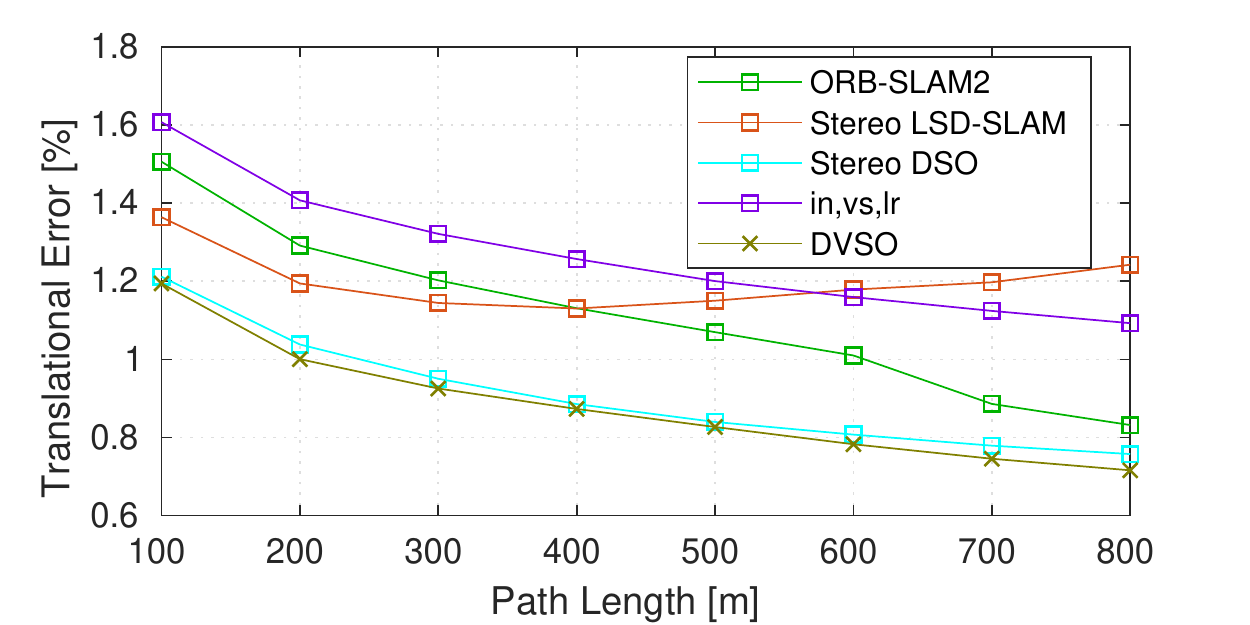}
	\end{subfigure}
	\begin{subfigure}[t]{.49\textwidth}
		\centering
		\includegraphics[width=\textwidth]{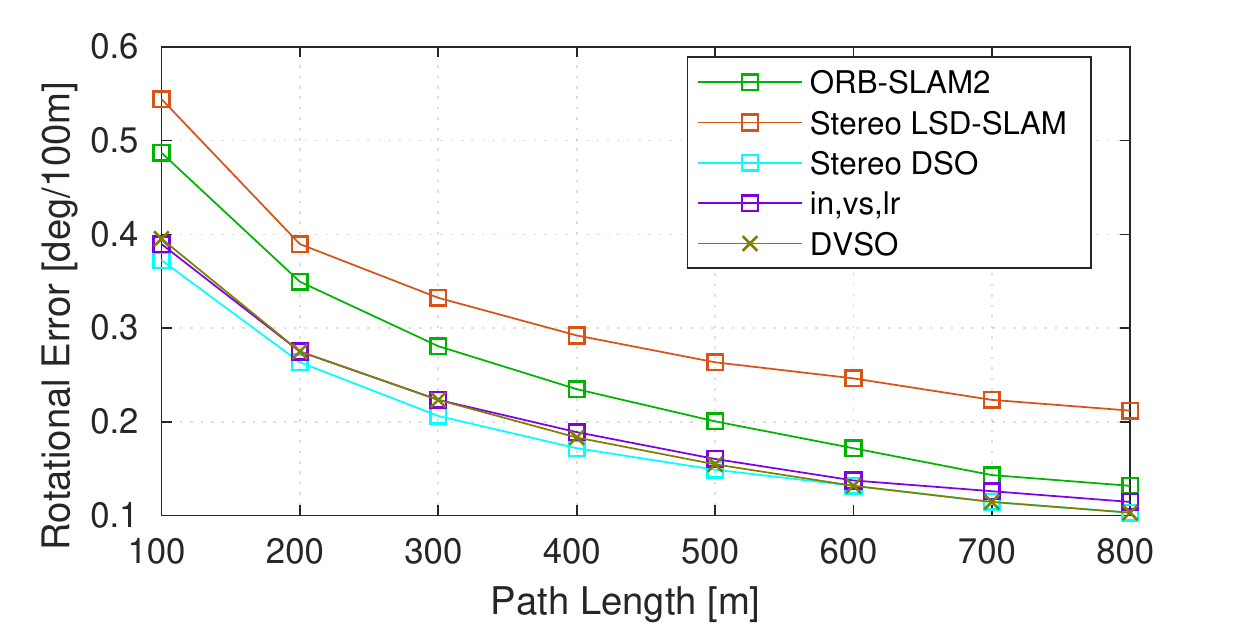}
	\end{subfigure}
	\caption{Evaluation results on the KITTI odometry test set. 
		We show translational and rotational errors with respect to 
		path length intervals. For translational errors, DVSO achieves 
		comparable performance to Stereo LSD-SLAM, while for rotational errors, DVSO 
		achieves comparable results to Stereo DSO 
		and better results than all other methods. Note that with virtual baseline 
		tuning, DVSO achieves the best performance among all the 
		methods evaluated.}
	\label{fig:kitti_test_set}
\end{figure}

We also compare DVSO with other state-of-the-art \emph{stereo} visual 
odometry systems on the sequences 00-10. The sequences with marker $^*$ are 
used for training StackNet and the sequences with marker $^\dagger$ are not 
used for training the network. In~\Cref{tab:compare_other_stereo} and 
the following tables, DVSO means our full approach with baseline tuning ($in,vs,lr,tb$). 
The average RMSE of DVSO without baseline tuning is better than Stereo 
LSD-VO, but not as good as Stereo DSO~\cite{wang2017stereoDSO} or 
ORB-SLAM2~\cite{mur2017orb} (stereo, without global optimization and loop closure).
Importantly, DVSO uses only monocular images. With the baseline tuning, 
DVSO achieves even better average performance than all other stereo systems 
on both rotational and translational errors.~\Cref{fig:kitti_traj} shows the 
estimated trajectory on sequence 00. Both monocular 
ORB-SLAM2 and DSO suffer from strong scale drift, while DVSO achieves 
superior performance on eliminating the scale drift. We also show the estimated 
trajectory on 00 by running DVSO using the 
depth map predicted by Godard et 
al.~\cite{godard2016unsupervised} with the model trained on the Cityscapes 
and the KITTI dataset. For the results in~\Cref{fig:kitti_traj} 
our depth predictions are more accurate.~\Cref{fig:kitti_test_set} shows 
the evaluation result of the sequences 11-21 by submitting results of DVSO with and without baseline tuning to the KITTI odometry 
benchmark.
Note that 
in~\Cref{fig:kitti_test_set}, Stereo LSD-SLAM and ORB-SLAM2 are both full 
stereo SLAM approaches with global optimization and loop closure. 
For qualitative comparisons of further 
estimated trajectories, please refer to our 
\href{https://vision.in.tum.de/_media/spezial/bib/yang2018dvso-supp.pdf}{supplementary
	material}.

\newcolumntype{C}[1]{>{\centering\let\newline\\\arraybackslash\hspace{0pt}}m{#1}}
\begin{table}[tb]
	\centering
	\tiny
	\def\arraystretch{1.05}
	\setlength{\tabcolsep}{0em}
	\begin{tabular}{C{1.5cm}|C{0.9cm}C{0.9cm}|C{0.95cm}C{0.95cm}|C{0.75cm}C{0.75cm}|C{0.75cm}C{0.75cm}|C{2pt}|C{0.8cm}C{0.8cm}}
		& \multicolumn{2}{c|}{St. LSD-VO~\cite{engel2015large}} & 
		\multicolumn{2}{c|}{ORB-SLAM2~\cite{mur2017orb}} 
		&\multicolumn{2}{c|}{St. 
			DSO~\cite{wang2017stereoDSO}} & 
		\multicolumn{2}{c|}{$in,vs,lr$}&&\multicolumn{2}{c}{DVSO}\\\cline{1-9}
		\cline{11-12}
		Seq. & $t_{rel}$     & $r_{rel}$ & $t_{rel}$     & $r_{rel}$    & 
		$t_{rel}$     & $r_{rel}$    & 
		$t_{rel}$      & $r_{rel}$&&$t_{rel}$ & $r_{rel}$\\
		00$^\dagger$   & 1.09 & 0.42
		& \textit{0.83} & 0.29 & 0.84 & \textit{0.26} & 0.93 & 
		\textbf{0.24}&&\textbf{0.71} & 
		\textbf{0.24}\\
		03$^\dagger$   & 1.16 & 0.32
		& \textbf{0.71} & \textit{0.17} & 0.92 & \textbf{0.16} 
		&2.56&0.18&&\textit{0.77} & 
		0.18\\
		04$^\dagger$   & \textit{0.42} & 0.34
		& 0.45 & 0.18 & 0.65 & 0.15 &0.67&\textit{0.07}&&\textbf{0.35} & 
		\textbf{0.06}\\
		05$^\dagger$   & 0.90 & 0.34
		& \textit{0.64} & 0.26 &0.68 & \textbf{0.19} & 
		\textit{0.64}&0.23&&\textbf{0.58} & 
		\textit{0.22}\\
		07$^\dagger$   & 1.25 & 0.79
		& \textit{0.78} & 0.42 & 0.83 & \textit{0.36} 
		&0.80&0.38&&\textbf{0.73} & 
		\textbf{0.35}\\\cline{1-9}
		\cline{11-12}
		01$^\ast$   & 2.13 & 0.37 
		& \textbf{1.38} & 0.20 &1.43 & \textbf{0.09} & 
		1.52&0.12&&\textbf{1.18} & 
		\textit{0.11}\\
		02$^\ast$   & 1.09 & 0.37 
		& \textit{0.81} & 0.28 &\textbf{0.78} & \textbf{0.21} & 
		1.05&0.23&&0.84 & 
		\textit{0.22}\\
		06$^\ast$   & 1.28 & 0.43
		& 0.82 & 0.25 & \textbf{0.67} & \textbf{0.20} 
		&0.80&\textit{0.24}&&\textit{0.71} & 
		\textbf{0.20}\\
		08$^\ast$   & 1.24 & 0.38
		& 1.07 & 0.31 & \textbf{0.98} & \textbf{0.25} 
		&1.10&\textit{0.26}&&\textbf{1.03} & 
		\textbf{0.25}\\
		09$^\ast$   & 1.22 & 0.28
		& \textbf{0.82} & 0.25 & 0.98 & \textbf{0.18} 
		&0.95&\textit{0.21}&&\textit{0.83} & 
		\textit{0.21}\\
		10$^\ast$  & 0.75 & 0.34 
		& \textit{0.58} & 0.28 & \textbf{0.49} & \textbf{0.18} 
		&0.59&0.22&&0.74 & 
		\textit{0.21}\\
		\hline
		\hline
		mean$^\dagger$ & 0.96 & 0.44& \textit{0.68} & 0.26 &0.78 & 
		\textit{0.22} & 
		1.12&\textit{0.22}&&\textbf{0.63} & \textbf{0.21}\\
		mean$^\ast$ & 1.29 & 0.36& \textit{0.91} & 0.26 &\textbf{0.89} & 
		\textbf{0.19} & 
		1.00&0.21&&\textbf{0.89} & \textit{0.20}\\
		overall mean & 1.14 & 0.40& \textbf{0.81} & 0.26 &0.84 & 
		\textbf{0.20} & 
		1.06&\textit{0.22}&&\textbf{0.77} & \textbf{0.20}\\
	\end{tabular}
	
	\caption{Comparison with state-of-the-art stereo visual odometry. 
		DVSO: our full approach ($in,vs,lr,tb$). Global optimization and loop-closure are turned off for 
		stereo ORB-SLAM2 and Stereo LSD-SLAM. DVSO (monocular) achieves comparable 
		performance to 
		these 
		stereo methods.}
	\label{tab:compare_other_stereo}
\end{table}

We also compare DVSO with DeepVO~\cite{wang2017deepvo}, UnDeepVO~\cite{li2017undeepvo} 
and SfMLearner~\cite{zhou2017unsupervised} which are  
deep learning based visual odometry systems trained end-to-end on KITTI. As shown 
in~\Cref{tab:compare_deep}, on all available sequences, DVSO achieves better performance than the other two end-to-end 
approaches.~\Cref{tab:compare_deep} also shows the comparison 
with the deep learning based scale recovery methods for monocular VO proposed by 
Yin et al.~\cite{yin2017scale}. DVSO also outperforms their method.
In the 
\href{https://vision.in.tum.de/_media/spezial/bib/yang2018dvso-supp.pdf}{supplementary
	material}, we also show the estimated trajectory on the 
Cityscapes \textit{Frankfurt} sequence to demonstrate generalization 
capabilities.

\begin{table}[tb]
	\centering
	\tiny
	\setlength{\tabcolsep}{0em}
	\begin{tabular}{C{0.8cm}|C{0.8cm}C{0.8cm}|C{0.8cm}C{0.8cm}|C{0.8cm}C{0.8cm}|C{0.8cm}C{0.8cm}|C{0.75cm}C{0.75cm}|C{2pt}|C{0.8cm}C{0.8cm}}
		& \multicolumn{2}{c|}{DeepVO~\cite{wang2017deepvo}} & 
		\multicolumn{2}{c|}{UnDeepVO~\cite{li2017undeepvo}} & 
		\multicolumn{2}{c|}{Yin et al.~\cite{yin2017scale}} &
		\multicolumn{2}{c|}{SfMLearner~\cite{zhou2017unsupervised}} &    
		\multicolumn{2}{c|}{$in,vs,lr$}&&\multicolumn{2}{c}{DVSO}\\\cline{1-11}\cline{13-14}
		Seq. & $t_{rel}$ & $r_{rel}$ & $t_{rel}$ & $r_{rel}$ & $t_{rel}$ & $r_{rel}$ & 
		$t_{rel}$ & 
		$r_{rel}$ & $t_{rel}$ & 
		$r_{rel}$ && $t_{rel}$ & $r_{rel}$\\
		00$^\dagger$&$-$& $-$ & 4.41 & \textit{1.92}
		& $-$ & $-$ 
		& 66.35 & 6.13
		& \textit{0.93}&\textbf{0.24}&&\textbf{0.71} & \textbf{0.24}\\
		
		03$^\dagger$   & 8.49 & 6.89 & 5.00 & \textit{6.17}
		& $-$ & $-$ 
		& 10.78 & 3.92
		&\textit{2.56}&\textbf{0.18}&&\textbf{0.77} & \textbf{0.18}\\
		
		04$^\dagger$   & 7.19 & 6.97 & 4.49 & 2.13
		& $-$ & $-$  
		& 4.49 & 5.24
		&\textit{0.67}&\textit{0.07}&&\textbf{0.35} & \textbf{0.06}\\
		
		05$^\dagger$   &2.62 & 3.61 & 3.40 & 1.50
		& $-$ & $-$ 
		& 18.67 & 4.10
		& \textit{0.64}&\textit{0.23}&&\textbf{0.58} & \textbf{0.22}\\
		
		07$^\dagger$   &3.91 & 4.60 & 3.15 & 2.48
		& $-$ & $-$ 
		& 21.33 & 6.65
		& \textit{0.80}&\textit{0.38}&&\textbf{0.73} & 
		\textbf{0.35}\\\cline{1-11}\cline{13-14}
		
		01$^\ast$&$-$& $-$ & 69.07 & 1.60
		& $-$ & $-$ 
		& 35.17 & 2.74
		& \textit{1.52}&\textit{0.12}&&\textbf{1.18} & \textbf{0.11}\\
		
		02$^\ast$   & $-$ & $-$ & 5.58 & 2.44
		& $-$ & $-$ 
		& 58.75 & 3.58
		&\textit{1.05}&\textit{0.23}&&\textbf{0.84} & \textbf{0.22}\\
		
		06$^\ast$   &5.42 & 5.82 & 6.20 & 1.98
		& $-$ & $-$ 
		& 25.88 & 4.80
		& \textit{0.80}&\textit{0.24}&&\textbf{0.71} & \textbf{0.20}\\
		
		08$^\ast$   & $-$ & $-$ & 4.08 & 1.79
		& 2.22 & \textbf{0.10} 
		& 21.90 & 2.91
		&\textit{1.10}& 0.26&&\textbf{1.03} & \textit{0.25}\\
		
		09$^\ast$&$-$& $-$ & 7.01 & 3.61
		& 4.14 & \textbf{0.11} 
		&18.77 &3.21
		&0.95&\textit{0.21}&&\textbf{0.83} & \textit{0.21}\\
		
		10$^\ast$   & 8.11 & 8.83 & 10.63 & 4.65
		& 1.70 & \textbf{0.17} 
		&14.33 &3.30 
		&\textbf{0.59}&0.22&&\textit{0.74} & \textit{0.21}\\
		\hline
	\end{tabular}
	\caption{Comparison with deep learning approaches. Note that Deep 
		VO~\cite{wang2017deepvo} is trained on sequences 00, 02, 08 and 09 of 
		the KITTI Odometry Benchmark. UnDeepVO~\cite{li2017undeepvo} and 
		SfMLearner~\cite{zhou2017unsupervised} are trained unsupervised on 
		seqs 00-08 end-to-end. Results of DeepVO and UnDeepVO taken from~\cite{wang2017deepvo} and~\cite{li2017undeepvo} 
		while for SfMLearner we ran their 
		pre-trained model. Our DVSO clearly outperforms state-of-the-art deep learning 
		based VO methods.}
	\label{tab:compare_deep}
\end{table}

\section{Conclusion}\label{sec:conclusion}
We presented a novel monocular visual odometry system, DVSO, which recovers metric scale and 
reduces scale drift in geometric monocular VO. A deep learning approach predicts monocular
depth maps for the input images which are used to initialize sparse depths in DSO to a consistent 
metric scale. Odometry is further improved by a novel 
virtual stereo term that couples estimated depth in windowed bundle adjustment with the monocular depth predictions. 
For monocular depth prediction we have presented a semi-supervised deep learning 
approach, which utilizes a self-supervised image reconstruction loss and sparse depth predictions from Stereo DSO as 
ground truth depths for supervision. A stacked network
architecture predicts state-of-the-art refined disparity estimates. 

Our evaluation conducted on the KITTI odometry benchmark demonstrates that DVSO outperforms the 
state-of-the-art monocular methods by a large margin and 
achieves comparable results to stereo VO methods. With virtual baseline tuning, 
DVSO can even outperform state-of-the-art stereo VO 
methods, i.e., Stereo LSD-VO, ORB-SLAM2 without global optimization and loop closure, and Stereo DSO, while using 
only monocular images.

The key practical benefit of the proposed method is that it allows us to recover accurate and scale-consistent odometry with only a single camera.
Future work could comprise fine-tuning of the network inside the odometry pipeline end-to-end.
This could enable the system to adapt online to new scenes and camera setups.
Given that the deep net was trained on driving sequences, in future work we 
also plan to investigate how much the proposed approach can generalize to other 
camera trajectories and environments.

\subsubsection{Acknowledgements} We would like to thank Cl\'{e}ment Godard, 
Yevhen Kuznietsov, Ruihao Li and Tinghui Zhou for providing the data and code. 
We also would like to thank Martin Schw\"{o}rer for the fruitful 
discussion.

\end{document}